\documentclass[letterpaper, 10 pt, conference]{ieeeconf}  % Comment this line out if you need a4paper

\IEEEoverridecommandlockouts                              % This command is only needed if 
\usepackage[T1]{fontenc} % preserve the template Times fonts across TeX engines
\usepackage{graphicx}
\usepackage{booktabs}
\usepackage{amsmath,amssymb}
\usepackage{cite}   % compressed numeric citations, e.g. [1]-[3]
\usepackage{balance} % balance the final page after the experimental floats

\title{\LARGE \bf
EditWM: Event-Decomposed World Modeling with Incremental Correction for End-to-End Autonomous Driving
}

\author{
Junjie Yang$^{1}$,
Qingwei Zeng$^{2}$,
Youyou Li$^{2}$,
Zicheng Ding$^{2}$, \\
Ziyi Shi$^{1}$,
Shuqi Shen$^{3}$,
Hongliang Lu$^{2}$,
Hai Yang$^{1}$
\thanks{$^{1}$The Hong Kong University of Science and Technology}
\thanks{$^{2}$Southern University of Science and Technology}
\thanks{$^{3}$The Chinese University of Hong Kong, Shenzhen}
}
\begin{document}

\maketitle
\thispagestyle{empty}
\pagestyle{empty}
\raggedbottom

%%%%%%%%%%%%%%%%%%%%%%%%%%%%%%%%%%%%%%%%%%%%%%%%%%%%%%%%%%%%%%%%%%%%%%%%%%%%%%%%
\begin{abstract}
World models support autonomous driving by predicting the scene evolution associated with candidate trajectories. Driving dynamics differ in predictability, motivating a distinction between regular evolution and event-induced deviations that call for selective correction. We propose EditWM, a world model that decomposes future prediction into normal evolution and event-driven incremental correction in compact visual feature space. A trajectory-conditioned normal predictor provides the base forecast and is then frozen for correction learning. A correction decoder compares this forecast with observation history and planned actions, producing a bounded feature update whose contribution is regulated by a learned gate. The corrected future features condition trajectory scoring through candidate-specific cross-attention, linking world modeling to plan selection. At inference, EditWM uses only past and current observations, ego state, and candidate trajectories. Across all 12,146 NAVSIM navtest scenes, expert-trajectory-conditioned evaluation shows a 5.35\% reduction in future-feature MSE over Normal, with improvements in 83.54\% of scenes. The system achieves 91.05 EPDMS on a 100-point scale using the official EPDMS evaluator. These results demonstrate improved future-feature prediction and competitive trajectory selection when corrected future representations are integrated into planning.
\end{abstract}

%%%%%%%%%%%%%%%%%%%%%%%%%%%%%%%%%%%%%%%%%%%%%%%%%%%%%%%%%%%%%%%%%%%%%%%%%%%%%%%%
\section{INTRODUCTION}

End-to-end autonomous driving has emerged as a promising paradigm that maps sensor observations directly to future trajectories~\cite{hu2023uniad,jiang2023vad,chen2024e2esurvey,chitta2022transfuser,sun2025sparsedrive}. However, the quality of a candidate trajectory depends not only on the current scene, but also on how the surrounding environment will evolve if that trajectory is executed. World models~\cite{ha2018worldmodels,hafner2019planet} have therefore been introduced into autonomous driving to predict the future scene evolution associated with different candidate trajectories~\cite{wang2024drivewm,wote2025,gao2024vista}, providing planning with an anticipatory view of their potential consequences. These predictions can further support trajectory evaluation, allowing the planner to select among candidates according to the futures they are expected to produce~\cite{li2024hydramdp,ang2026clover}.

\begin{figure*}[t]
    \centering
    \parbox{\textwidth}{\centering
        \includegraphics[width=0.85\linewidth]{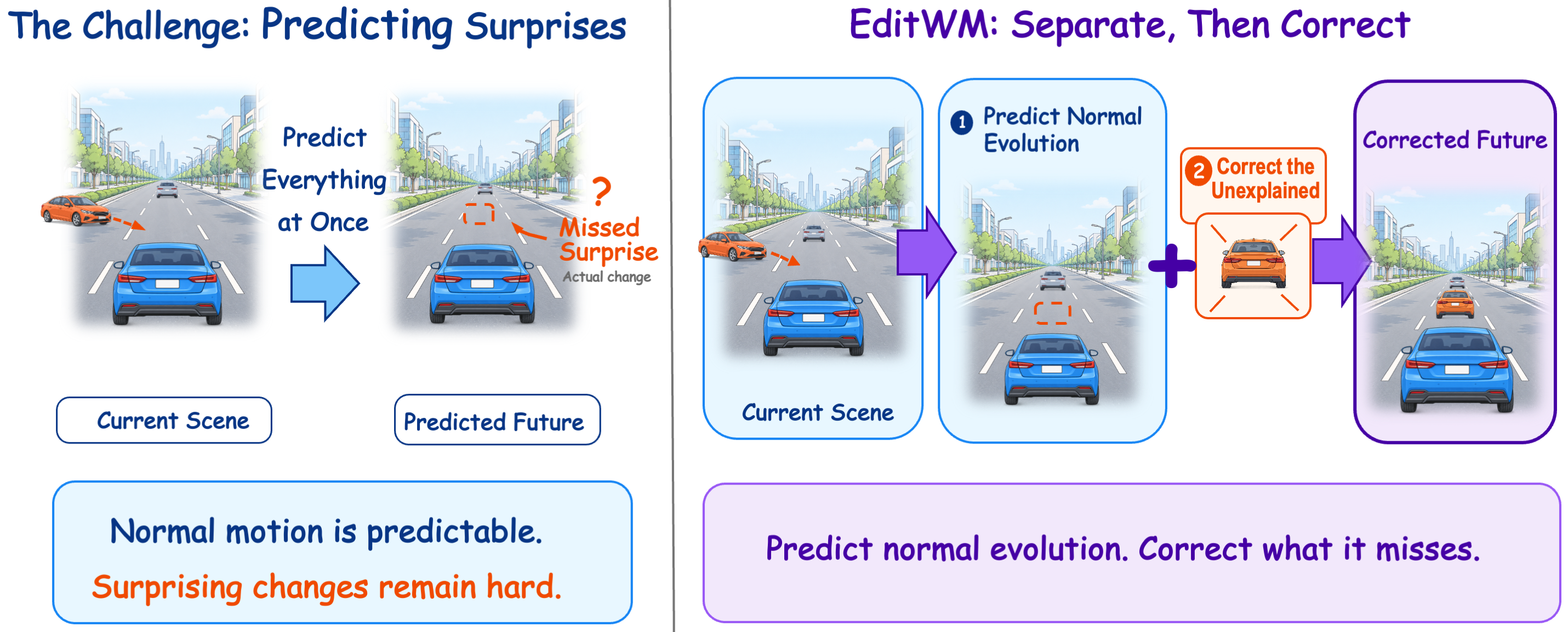}%
    }
    \caption{The core idea of EditWM: predict normal evolution and correct what it misses. Left: a cut-in illustrates feature changes left unexplained by a normal forecast. Right: EditWM retains the normal forecast and selectively refines it with a gated feature correction. Orange highlights the unexplained change and its correction. The road scenes schematically illustrate latent feature prediction and correction.}
    \label{fig:teaser}
\end{figure*}

Existing driving world models typically predict the future scene as a whole~\cite{wang2024drivewm,wote2025}. Visual world models generate future observations to preserve scene appearance and spatial structure~\cite{wang2024drivedreamer,gao2024vista}, whereas latent world models predict future scene representations conditioned on the ego trajectory or driving intent, avoiding explicit pixel reconstruction~\cite{li2025law,zheng2025world4drive}. Despite operating in different prediction spaces, both paradigms generally treat the complete future evolution as a unified prediction target.

% However, driving scenes combine regular evolution with unexpected traffic events. Events such as cut-ins, sudden braking, and the appearance of new agents can disrupt the prevailing scene evolution and have consequential effects on trajectory selection~\cite{min2024driveworld}. Such events can induce feature changes that a normal forecast does not fully capture. Fig.~\ref{fig:teaser} illustrates this motivation with a cut-in: the surrounding scene may evolve regularly while the entering vehicle introduces additional variation. The missing element is an explicit mechanism for correcting what the normal prediction fails to explain.
However, not all future changes in a driving scene are equally predictable~\cite{min2024driveworld}. Much of the scene follows regular evolution induced by ego motion, persistent geometry, and smooth agent dynamics, while a smaller but decision-critical fraction arises from abrupt or interaction-driven changes that are difficult to capture with nominal forecasting alone (Fig.~\ref{fig:teaser}, left). Existing world models typically ask a single predictor to model both regimes simultaneously~\cite{wang2024drivewm,wote2025}. This unified formulation overlooks their asymmetric predictability and provides no explicit mechanism to preserve reliable forecasts while selectively correcting the changes they fail to explain. We therefore ask: rather than repeatedly predicting the entire future from scratch, can a driving world model retain what is already predictable and focus its capacity on correcting what remains unexplained?

To address this asymmetry, we propose EditWM, a world model for end-to-end autonomous driving based on event-driven incremental correction. As illustrated on the right of Fig.~\ref{fig:teaser}, EditWM separates future prediction into normal evolution and surprise correction, preserving the normal forecast as the basis for refinement. This design focuses correction learning on feature changes left unexplained by the normal forecast, while reusing its prediction of regular evolution. The corrected future representations guide candidate trajectory evaluation, allowing planning decisions to account for anticipated scene evolution using only past and current observations and proposed actions.

Our contributions are threefold. First, motivated by heterogeneous predictability in driving dynamics, we introduce a decomposed future-modeling paradigm that preserves normal forecasts and selectively corrects their unexplained feature changes. Second, we connect this decomposition to planning through candidate-specific corrected futures: each candidate is evaluated using a normal forecast refined by its own learned surprise correction, making selective refinement part of the evidence used for trajectory selection. Third, we conduct a systematic evaluation on all 12,146 navtest scenarios. EditWM reduces the MSE of future-feature prediction by 5.35\% and yields more accurate future predictions in 83.54\% of the scenarios. For planning, the full system achieves an EPDMS of 91.05, demonstrating competitive performance relative to leading published methods.

\section{RELATED WORK}
\label{sec:related_work}

\subsection{World Models for Driving}

Driving world models connect scene prediction to the consequences of ego actions. Visual approaches establish this connection through future observations: DriveDreamer learns traffic-conditioned video generation~\cite{wang2024drivedreamer}, Drive-WM compares action-conditioned multiview futures for planning~\cite{wang2024drivewm}, and Vista extends visual forecasting with versatile action control~\cite{gao2024vista}. Their generated observations expose how a driving scene may evolve, while planning ultimately requires a representation that helps distinguish candidate actions.

Compact world models bring future prediction closer to the planner's internal representation. OccWorld models temporal evolution in occupancy space~\cite{zheng2024occworld}, while LAW uses ego-trajectory-conditioned future features as a self-supervised learning signal~\cite{li2025law}. World4Drive further combines intention-aware latent prediction with trajectory selection~\cite{zheng2025world4drive}, and WoTE evaluates candidate trajectories using their predicted BEV futures~\cite{wote2025}. These methods establish that predicted representations can inform planning directly. EditWM builds on this connection by explicitly distinguishing normal forecasting from selective correction, motivated by the heterogeneous predictability of driving dynamics.

\subsection{Structured Dynamics and Selective Updates}

Factoring scene evolution can reduce the burden of modeling all dynamics together. DriveWorld separates dynamic memory from static scene propagation to learn spatiotemporal representations for driving~\cite{min2024driveworld}. Such a division motivates structured modeling, but separating static and dynamic content does not by itself identify what a particular future forecast fails to explain.

A complementary direction regulates when latent states should change. GateL0RD encourages sparse latent updates using gates and an $L_0$ penalty~\cite{gumbsch2021gatel0rd}, whereas Variational Sparse Gating learns stochastic binary gates over latent feature updates~\cite{jain2022vsg}. EditWM applies selective updates to a separately trained normal forecast. Its gate controls a bounded correction that accounts for feature changes left unexplained by normal evolution. This connects structured future prediction with selective adjustment, allowing the model to preserve its normal forecast while incorporating additional evidence about unexpected changes.

\subsection{Candidate Generation and Trajectory Scoring}

Candidate-based planning first requires alternatives that cover plausible driving behaviors. VADv2 models a probabilistic distribution over driving actions~\cite{chen2024vadv2}, and DiffusionDrive generates diverse trajectories with truncated diffusion around multimodal anchors~\cite{liao2025diffusiondrive}. Once alternatives are available, the central question becomes how to rank them. Hydra-MDP distills imitation and planning-metric supervision into trajectory evaluation~\cite{li2024hydramdp}, while GTRS improves scoring across coarse trajectory vocabularies and fine-grained generated proposals~\cite{li2025gtrs}.

These approaches improve the available candidates and the supervision used to evaluate them. The complementary question addressed by EditWM is what future evidence should inform that evaluation. We retain a fixed candidate pool and metric-based scoring objective, and augment each candidate with its own corrected future features. This connects candidate-conditioned forecasting to trajectory ranking: the world model predicts the scene under a proposed action, and the scorer uses that prediction to assess the action's anticipated outcome.

\begin{figure*}[t]
    \centering
    \parbox{\textwidth}{\centering
        \includegraphics[width=0.9\linewidth]{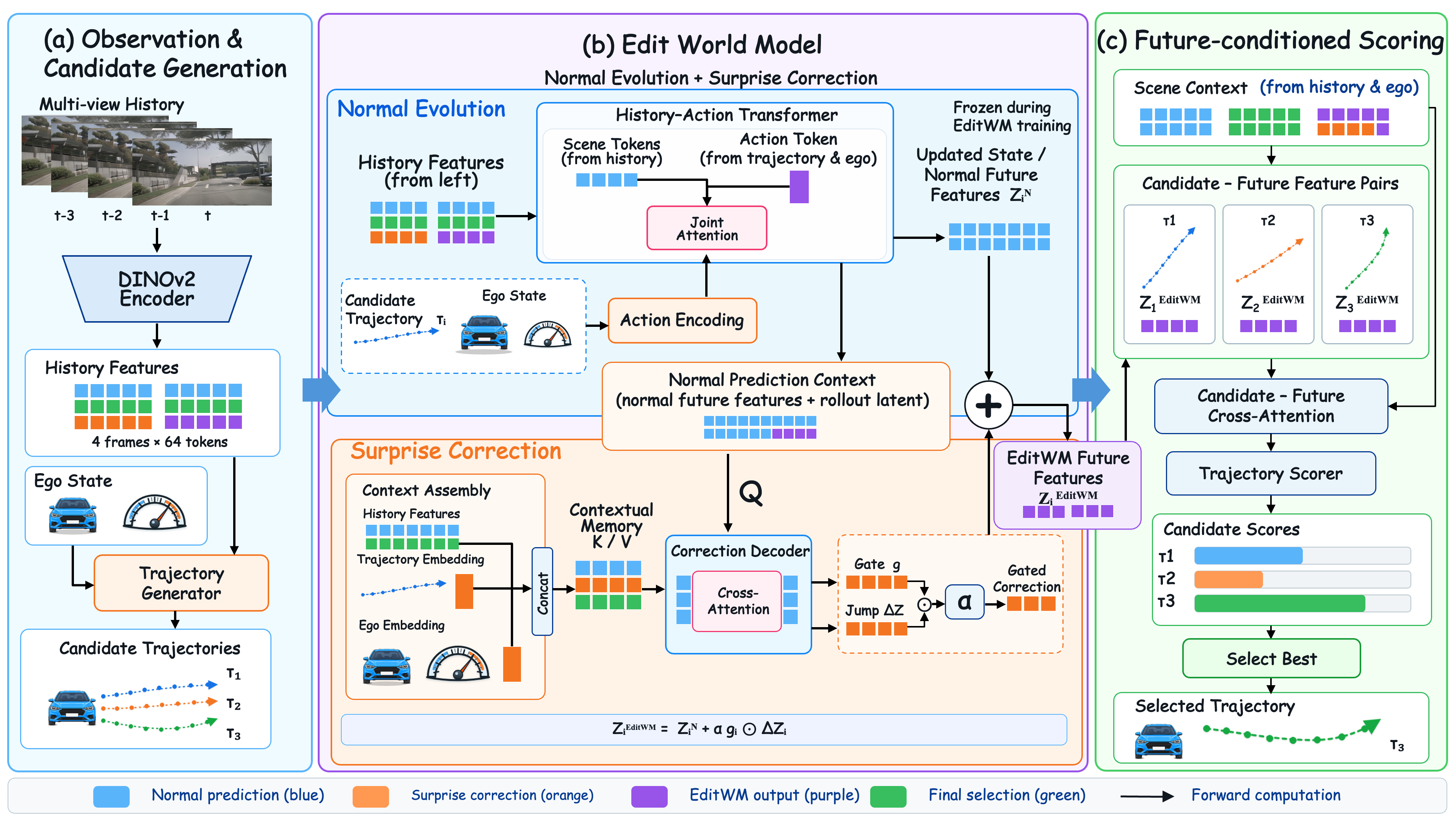}%
    }
    \caption{Overview of the EditWM world model and its integration with trajectory scoring. (a) Multi-view history is encoded with DINOv2, while a trajectory generator produces candidate plans. (b) For each candidate, a normal dynamics predictor produces future features, and a correction decoder predicts a gated residual from normal prediction context and history--action memory. The normal forecast and gated incremental correction together form the EditWM future representation. The frozen normal predictor supplies feature-residual supervision for correction and gating. (c) Each candidate and its corresponding predicted future features are fused for trajectory scoring and selection. World-model parameters are shared across candidates; colored token blocks depict latent features.}
    \label{fig:framework}
\end{figure*}

\section{METHOD}
\label{sec:method}

\subsection{Problem Formulation and Overview}
\label{sec:method_problem}

At time $t$, a frozen DINOv2-based encoder~\cite{oquab2023dinov2} maps multi-view images to history $\mathcal H_t=\{Z_{t-L+1},\ldots,Z_t\}$. Each of the $L$ frames contains $N$ visual tokens of dimension $d$, giving $Z\in\mathbb R^{N\times d}$. The planner also receives ego state $e_t$ and candidates $\mathcal C_t=\{\tau_i\}_{i=1}^{K}$, where $K$ is the candidate count and $\tau_i\in\mathbb R^{H\times3}$ contains $H$ time-indexed planar positions and headings in the current ego frame. As shown in Fig.~\ref{fig:framework}, EditWM predicts a terminal future representation for each candidate through normal evolution and incremental correction, then uses it to guide trajectory selection. World-model parameters are shared across candidates, while their action conditions distinguish the futures supplied to the scorer.

\subsection{Normal Evolution as a Prediction Reference}
\label{sec:normal_dynamics}

Following WoTE's state--action Transformer design~\cite{wote2025}, a history encoder initializes the shared scene state $S_0\in\mathbb R^{N\times d}$, while trajectory and ego embeddings initialize candidate action $a_{i,0}\in\mathbb R^d$. The history encoder aggregates observed frames with camera, slot, and time information. The action embedding summarizes the proposed motion together with the ego state, giving the dynamics a candidate-specific condition throughout prediction. Starting from $S_{i,0}=S_0$, normal evolution is
\begin{equation}
\begin{aligned}
[a_{i,m+1};S_{i,m+1}]
 &= D_\theta([\tilde a_{i,m};\tilde S_{i,m}]),\\
\hat Z_i^N &= Z_t+P_\theta(S_{i,M}).
\end{aligned}
\label{eq:normal_rollout}
\end{equation}
Here $D_\theta$ is a shared Transformer with normal-predictor parameters $\theta$, $m=0,\ldots,M-1$ indexes $M$ transitions, and $[\,;\,]$ concatenates tokens. The conditioned action $\tilde a_{i,m}$ adds the pooled embeddings of the next two waypoints, transition time, and action type; $\tilde S_{i,m}$ adds camera, slot, and transition-time embeddings to the scene state. Projection $P_\theta$ maps the terminal state to a change from current features $Z_t$, yielding the normal future $\hat Z_i^N$. This expresses the normal forecast in the same visual feature space as the observations, providing a reusable base for incremental correction.

After fitting Normal, we freeze it and define surprise as its unexplained feature residual:
\begin{equation}
R=Z^+-\hat Z^N(\mathcal H_t,e_t,\tau^{\mathrm{gt}}).
\label{eq:surprise_target}
\end{equation}
Here $Z^+\in\mathbb R^{N\times d}$ is the observed terminal future under expert trajectory $\tau^{\mathrm{gt}}$. Residual $R$ supervises the feature correction and provides the signal for learning its gate. Normal supplies the reference forecast, while EditWM learns to refine it using history--action context. At inference, the decoder predicts the correction and gate from the normal forecast and this context.

\subsection{Event-Driven Incremental Correction}
\label{sec:edit_correction}

After the normal rollout, the correction decoder refines its terminal forecast using historical and action evidence:
\begin{equation}
\begin{aligned}
Q_i &= \operatorname{LN}(\hat Z_i^N)+P_S(S_{i,M})+B_Q,\\
\mathcal M_i &= [E_h(\mathcal H_t);E_\tau(\tau_i);E_e(e_t)],\\
U_i &= \operatorname{Decoder}_\phi(Q_i,\mathcal M_i).
\end{aligned}
\label{eq:correction_decoder}
\end{equation}
The query $Q_i$ and decoder output $U_i$ have shape $N\times d$. Layer normalization $\operatorname{LN}$, rollout-state projection $P_S$, and positional/terminal-time embedding $B_Q$ anchor the query to the predicted future. The embeddings $E_h$, $E_\tau$, and $E_e$ produce $LN$ history tokens, $H$ ordered waypoint tokens, and one ego token, respectively, forming $\mathcal M_i\in\mathbb R^{(LN+H+1)\times d}$. History retains camera, slot, and time information; waypoint embeddings preserve pose and timing. The Transformer decoder, parameterized by $\phi$, uses this memory as keys and values, allowing the forecast to query evidence for its correction.

The normal future specifies the representation to be edited, while the terminal rollout state supplies its trajectory-conditioned dynamics context. History and waypoints provide evidence for the adjustment. Correction embeddings are learned separately from the frozen predictor, allowing the evidence representation to adapt around a fixed reference.

The decoder updates its future-token queries through self-attention, cross-attention to context memory, and a feed-forward transformation. Self-attention permits coordinated corrections across visual slots, while cross-attention conditions them on observed history and candidate actions.

Two heads predict a gate and bounded jump, whose product edits the normal future:
\begin{equation}
\begin{aligned}
g_i &= \sigma(P_g(U_i)),\\
\Delta Z_i &= B_\Delta\odot\tanh(P_J(U_i)),\\
\hat Z_i^{\mathrm{EditWM}} &= \hat Z_i^N+\alpha\,g_i\odot\Delta Z_i.
\end{aligned}
\label{eq:edit_decomposition}
\end{equation}
The linear heads $P_g$ and $P_J$ produce token-wise gate logits and jump features. Sigmoid $\sigma$ gives $g_i\in(0,1)^N$, while the hyperbolic tangent bounds $\Delta Z_i\in\mathbb R^{N\times d}$ using a positive fitting-set scale $B_\Delta$. The operator $\odot$ denotes element-wise multiplication, broadcasting gates across channels, and $\alpha\in[0,1]$ is a validation-selected global gain. The gate controls the local contribution of each jump; a vanishing gate recovers Normal. The update in Eq.~\ref{eq:edit_decomposition} preserves the normal forecast $\hat Z_i^N$ as its base and adds the selective correction $\alpha g_i\odot\Delta Z_i$ before trajectory scoring. The bounded increment preserves the base wherever the gated correction is small.

\subsection{Scoring with Candidate-Specific Future Features}
\label{sec:future_scoring}

To connect future prediction to planning, candidate features $c_i\in\mathbb R^d$ encode trajectory $\tau_i$, current scene $Z_t$, and ego state $e_t$, with interactions among candidates. Candidate-specific prediction maintains a separate future representation for each proposed motion. This gives the scorer action-dependent future context in addition to the shared evidence about the currently observed scene. Each candidate's future memory combines its corrected scene and change from the current observation:
\begin{equation}
\mathcal V_i=\operatorname{LN}(\hat Z_i^{\mathrm{EditWM}})
 +P_\delta(\hat Z_i^{\mathrm{EditWM}}-Z_t)+B_V.
\label{eq:future_memory}
\end{equation}
Here $\mathcal V_i\in\mathbb R^{N\times d}$ is the memory, $P_\delta$ projects the feature change, and $B_V$ supplies camera and slot embeddings. The normalized features expose predicted scene content, while the projected difference explicitly represents its evolution from the current state. Candidate--future attention retrieves relevant information from both through the following update:
\begin{equation}
\begin{aligned}
v_i &= \operatorname{CrossAttn}(\operatorname{LN}(c_i),\mathcal V_i,\mathcal V_i),\\
\tilde c_i &= c_i+P_f\bigl(v_i+\operatorname{FFN}(v_i)\bigr),\\
\ell_i &= h_\psi(\tau_i,\tilde c_i).
\end{aligned}
\label{eq:future_fusion}
\end{equation}
The attention takes query, key, and value in that order, retrieving $v_i\in\mathbb R^d$. A feed-forward network $\operatorname{FFN}$ and projection $P_f$ yield the fused candidate feature $\tilde c_i$ through a residual update. The scorer $h_\psi$, with parameters $\psi$, outputs nine component logits $\ell_i\in\mathbb R^9$. Each candidate attends only to its own future, and zero-initializing $P_f$ preserves the pretrained scorer's initial behavior. The residual connection retains the candidate's current-scene representation while learning how future evidence should modify it. Candidate interactions remain in the base encoder; future attention maintains the correspondence between each trajectory and its own prediction.

Sigmoid-activated logits are combined into score $s_i$ using the metric composition in Sec.~\ref{sec:experimental_setup}; the selected trajectory is $\tau^\star=\arg\max_{\tau_i\in\mathcal C_t}s_i$.

\subsection{Staged Learning of Prediction and Scoring}
\label{sec:method_training}

We train the modules in sequence so that correction learns against an established normal forecast and scoring subsequently adapts to a fixed world model.

\paragraph{Normal prediction}
We first train $\theta$ on expert trajectories using $\mathcal L_{\mathrm{feat}}(\hat Z^N,Z^+)$, defined as mean squared error plus $\lambda_{\mathrm{cos}}$ times mean token-wise cosine distance. Targets $Z^+$ come from the frozen visual encoder. We select a validation checkpoint, freeze Normal, and compute residuals for correction training.

\paragraph{Gated correction}
Suppressing the expert-trajectory index, token $n$ has residual energy $\rho_n=\|R_n\|_2^2/d$, where $R_n$ is the corresponding row of $R$. Its soft gate target is $y_n=\rho_n/(\rho_n+\kappa_n)$, with positive scale $\kappa_n$ calibrated on fitting data. The target increases with unexplained error relative to this scale, guiding where correction should contribute. We train with unit global gain:
\begin{equation}
\begin{aligned}
\tilde Z &= \hat Z^N+\operatorname{sg}(g)\odot\Delta Z,\\
\mathcal L_{\mathrm{EditWM}}
 &= \mathcal L_{\mathrm{feat}}(\tilde Z,Z^+)\\
 &\quad+\lambda_g\operatorname{BCE}(g,y)
 +\lambda_m\operatorname{mean}(g).
\end{aligned}
\label{eq:edit_training}
\end{equation}
Here $\operatorname{sg}$ stops gradients through gate values, and $\tilde Z$ is the training prediction. Binary cross-entropy $\operatorname{BCE}$ and $\operatorname{mean}$ average over tokens; $\lambda_g$ and $\lambda_m$ weight gate supervision and activation regularization. The feature loss trains the jump path and shared decoder, while the gate head learns through the gate-specific terms. Stopping the feature-loss gradient through gate values prevents reconstruction from directly increasing gate activation to reduce its error, while preserving the gated forward prediction. Validation selects the correction checkpoint and inference gain $\alpha$. This gain is then held fixed during scorer adaptation and deployment, so the scorer learns from the same corrected representation that it will receive at inference.

\paragraph{Trajectory scoring}
Finally, we freeze the world model and train the scorer and future-attention branch on the fixed candidate pool. Binary cross-entropy supervises the nine component predictions, averaging over available labels and masking missing entries. Each candidate is supervised by its own component labels, preserving the correspondence between the proposed trajectory, its predicted future, and the resulting quality targets. Perception and generation remain frozen. This stage learns how predicted future evidence should affect ranking while keeping the forecasting modules unchanged. Real future features supervise world-model learning, whereas candidate-quality labels supervise scoring; the deployed planner receives only observed history, ego state, and candidate trajectories. Hyperparameters and calibration details are given in Sec.~\ref{sec:experimental_setup}.

\section{EXPERIMENTS}
\label{sec:experiments}
% Declare the main-results float early so it stays with the experiment section.
\begin{table*}[t]
\caption{EPDMS leaderboard and component scores (0--100; higher is better).}
\label{tab:leaderboard}
\centering
\footnotesize
\setlength{\tabcolsep}{3pt}
\renewcommand{\arraystretch}{1.13}
\resizebox{\textwidth}{!}{%
\begin{tabular}{llrrrrrrrrrr}
\toprule
Method & Backbone & NC & DAC & DDC & TL & EP & TTC & LK & HC & EC & EPDMS \\
\midrule
Hydra-MDP++~\cite{li2025hydramdppp} & ResNet-34 & 98.4 & 98.0 & 99.4 & 99.8 & 87.5 & 97.7 & 95.3 & \textbf{98.3} & 77.4 & 85.1$^{*}$ \\
DiffusionDrive~\cite{liao2025diffusiondrive} & ResNet-34 & 98.2 & 96.2 & 99.5 & 99.8 & 87.4 & 97.4 & \textbf{97.0} & \textbf{98.3} & 87.8 & 84.8$^{*}$ \\
WoTE~\cite{wote2025} & ResNet-34 & 98.5 & 96.8 & 98.8 & 99.8 & 86.1 & 97.9 & 95.5 & \textbf{98.3} & 82.9 & 84.2 \\
iPad~\cite{niu2025ipad} & ResNet-34 & 98.7 & 97.8 & 99.1 & 99.8 & 84.0 & 98.0 & 96.0 & 98.0 & 68.2 & 84.1 \\
DriveSuprim~\cite{yao2026drivesuprim} & ResNet-34 & 97.8 & 97.9 & 99.5 & \textbf{99.9} & 90.6 & 97.1 & 96.6 & \textbf{98.3} & 77.9 & 86.0$^{*}$ \\
DiffusionDriveV2~\cite{zou2025diffusiondrivev2} & ResNet-34 & 97.7 & 96.6 & 99.2 & 99.8 & 88.9 & 97.2 & 96.0 & 97.8 & 91.0 & 87.5 \\
DrivoR~\cite{kirby2026drivor} & DINOv2-S & 99.2 & 98.7 & \textbf{99.6} & 99.8 & 79.8 & 98.7 & 94.5 & \textbf{98.3} & 80.6 & 88.40 \\
SparseDriveV2~\cite{sun2026sparsedrivev2} & ResNet-34 & 98.1 & 98.1 & \textbf{99.6} & 99.8 & \textbf{91.1} & 97.3 & 96.9 & 98.2 & 78.4 & 90.1 \\
CLOVER~\cite{ang2026clover} & DINOv2-S & \textbf{99.4} & \textbf{99.3} & 99.5 & 99.8 & 86.9 & \textbf{98.9} & 95.2 & 98.2 & 75.5 & 90.4 \\
\midrule
\textbf{Ours} & DINOv2-S & 97.8 & 98.8 & \textbf{99.6} & \textbf{99.9} & 87.5 & 96.9 & 94.8 & \textbf{98.3} & \textbf{96.6} & \textbf{91.05} \\
\bottomrule
\end{tabular}
}
\par\vspace{3pt}
\parbox{\linewidth}{\footnotesize $^{*}$ Original v2 evaluator before human-behavior filtering correction. Baseline results are taken from their reported evaluations. Best displayed values, including ties, are bold.}
\end{table*}

\subsection{Experimental Setup}
\label{sec:experimental_setup}

\paragraph{Data and representation}
We use NAVSIM~\cite{dauner2024navsim}, training the trajectory scorer on all 103,288 navtrain scenes and evaluating on all 12,146 navtest scenes. World-model training uses the 43,849 navtrain scenes with real future images, divided by source log into 39,337 fitting and 4,512 validation scenes. These world-model splits and navtest do not share physical frames. The DINOv2 representation contains 64 learned slots with 256 channels, with 16 slots per camera across four views. Normal conditions on four observation frames and an ego trajectory with eight waypoints over a nominal four-second horizon. Four rollout transitions each consume two successive waypoints.

\paragraph{Training and inference}
We use a fixed generator with 96 candidates per scene. Normal and EditWM each train for 30 epochs using AdamW, a learning rate of $2\times10^{-4}$, weight decay 0.01, and global batch size 64 across four GPUs. Normal is frozen during joint training of the Gate and Jump branches. We use cosine-loss weight $\lambda_{\mathrm{cos}}=0.1$, gate-supervision weight $\lambda_g=0.02$, and gate-activation weight $\lambda_m=0.001$. The gate energy scale $\kappa_n$ is the 80th percentile of fitting-set residual energy for token $n$; the jump bound $B_\Delta$ is four times the token-and-channel-wise residual root mean square on the same split. Validation selects Normal at epoch 29 and EditWM at epoch 30, with correction scale $\alpha=0.74884$. The scorer is initialized from pretrained weights and adapted for 24 epochs on all navtrain scenes and all 96 candidates. Perception, the generator, and the world model remain frozen during scorer training. Scorer evaluation uses the fixed final epoch. Future images provide training supervision and targets for feature-error measurement.

\paragraph{Evaluation protocol}
We report EPDMS~\cite{cao2025pseudosimulation} and nine components on a 0--100 scale: no-at-fault collisions (NC), drivable-area compliance (DAC), driving-direction compliance (DDC), traffic-light compliance (TL), ego progress (EP), time-to-collision compliance (TTC), lane keeping (LK), history comfort (HC), and extended comfort (EC). For candidate ranking, the product of predicted NC, DAC, DDC, and TL multiplies the weighted mean of predicted TTC, EP, LK, HC, and EC, with respective weights $(5,5,2,2,2)$. Planning metrics are computed with the official EPDMS evaluator for the selected trajectories. The reported EPDMS is the mean of per-scene total scores. External methods retain their previously reported evaluator protocols. For world-model evaluation, both predictors receive the expert trajectory and are compared against real terminal future features using MSE and cosine similarity.

\subsection{Overall Planning Performance}
\label{sec:planning_results}

Table~\ref{tab:leaderboard} compares the full system with nine representative planners using their reported EPDMS and component scores. Ours achieves 91.05 EPDMS on all 12,146 navtest scenes, the highest total among the listed methods. It exceeds SparseDriveV2's 90.1 by 0.95 points and DrivoR's 88.40 by 2.65 points.

Ours achieves the highest extended comfort score (EC=96.6 versus 91.0 for DiffusionDriveV2) and matches the best driving-direction compliance (DDC=99.6), traffic-light compliance (TL=99.9), and history comfort (HC=98.3). Together with DAC=98.8 and EP=87.5, these results highlight strong comfort and compliance with continued forward progress.

\subsection{Future Features and Trajectory Selection}
\label{sec:future_ablation}

Table~\ref{tab:future_ablation} compares planning performance on the same 96-candidate pool, scene indices, and labels. All three variants use the same general-purpose nine-head scorer architecture: each head predicts one EPDMS component, and the component predictions are combined using the scoring rule in Sec.~\ref{sec:experimental_setup}. The no-future variant uses only current features, Normal-only incorporates normal future features, and Ours incorporates EditWM-corrected future features. The Oracle selects the candidate with the highest total EPDMS in each scene; its component scores are those of the selected candidate.

\begin{table*}[t]
\caption{Future-feature and trajectory-selection comparison on the same K=96 candidate pool (0--100; higher is better).}
\label{tab:future_ablation}
\centering
\footnotesize
\setlength{\tabcolsep}{3pt}
\renewcommand{\arraystretch}{1.13}
\resizebox{\textwidth}{!}{%
\begin{tabular}{lrrrrrrrrrr}
\toprule
Method & NC & DAC & DDC & TL & EP & TTC & LK & HC & EC & EPDMS \\
\midrule
w/o Future Features & 97.55 & 98.67 & 99.59 & 99.70 & 87.34 & 96.80 & 94.75 & 98.21 & 96.36 & 90.84 \\
w/o EditWM (Normal-only) & 97.34 & 98.53 & 99.53 & 99.72 & 87.08 & 96.65 & 94.71 & 98.20 & 96.36 & 90.51 \\
\midrule
\textbf{Ours} & \textbf{97.8} & \textbf{98.8} & \textbf{99.6} & \textbf{99.9} & \textbf{87.5} & \textbf{96.9} & \textbf{94.8} & \textbf{98.3} & \textbf{96.6} & \textbf{91.05} \\
\midrule
Oracle & 99.79 & 99.69 & 99.79 & 99.97 & 95.56 & 99.77 & 99.16 & 98.35 & 97.99 & 97.73 \\
\bottomrule
\end{tabular}
}
\par\vspace{3pt}
\parbox{\linewidth}{\footnotesize All three variants use the same general-purpose nine-head scorer architecture. The no-future row uses only current features. Bold indicates the best displayed value excluding Oracle. Oracle reports the components of the total-score-maximizing candidate.}
\end{table*}

% Declare the prediction figure before its discussion to keep it on page six.
\begin{figure*}[t]
    \centering
    \parbox{\textwidth}{\centering
        \includegraphics[width=\linewidth]{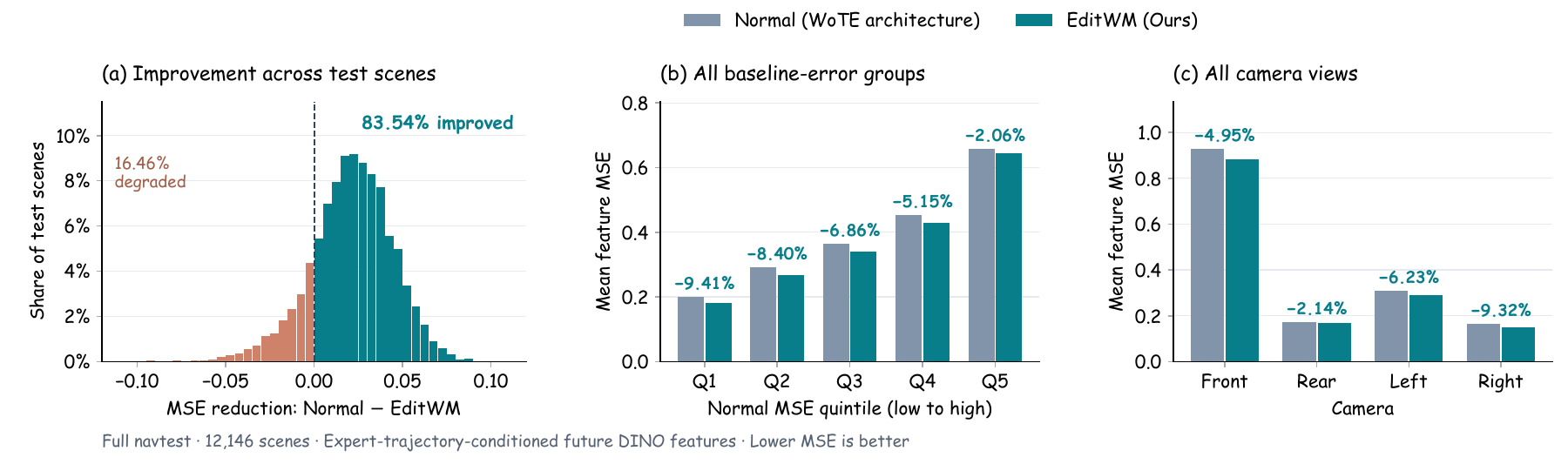}%
    }
    \caption{Future DINO feature prediction on all 12,146 navtest scenes, conditioned on the expert trajectory. (a) Per-scene MSE reduction, defined as Normal minus EditWM; negative values indicate degradation and are retained. (b) Mean MSE for five equally sized groups ordered by Normal's prediction error. (c) Mean MSE for each camera view. Percentage annotations in (b) and (c) give the relative reduction in group-mean MSE. EditWM improves 83.54\% of scenes and reduces overall MSE by 5.35\%.}
    \label{fig:prediction_quality}
\end{figure*}

% A borderless text box follows the template's graphic-container suggestion.
% figure* spans both columns; the caption remains below the embedded vector PDF.
\begin{figure*}[t]
    \centering
    \parbox{\textwidth}{\centering
        \includegraphics[width=0.85\linewidth]{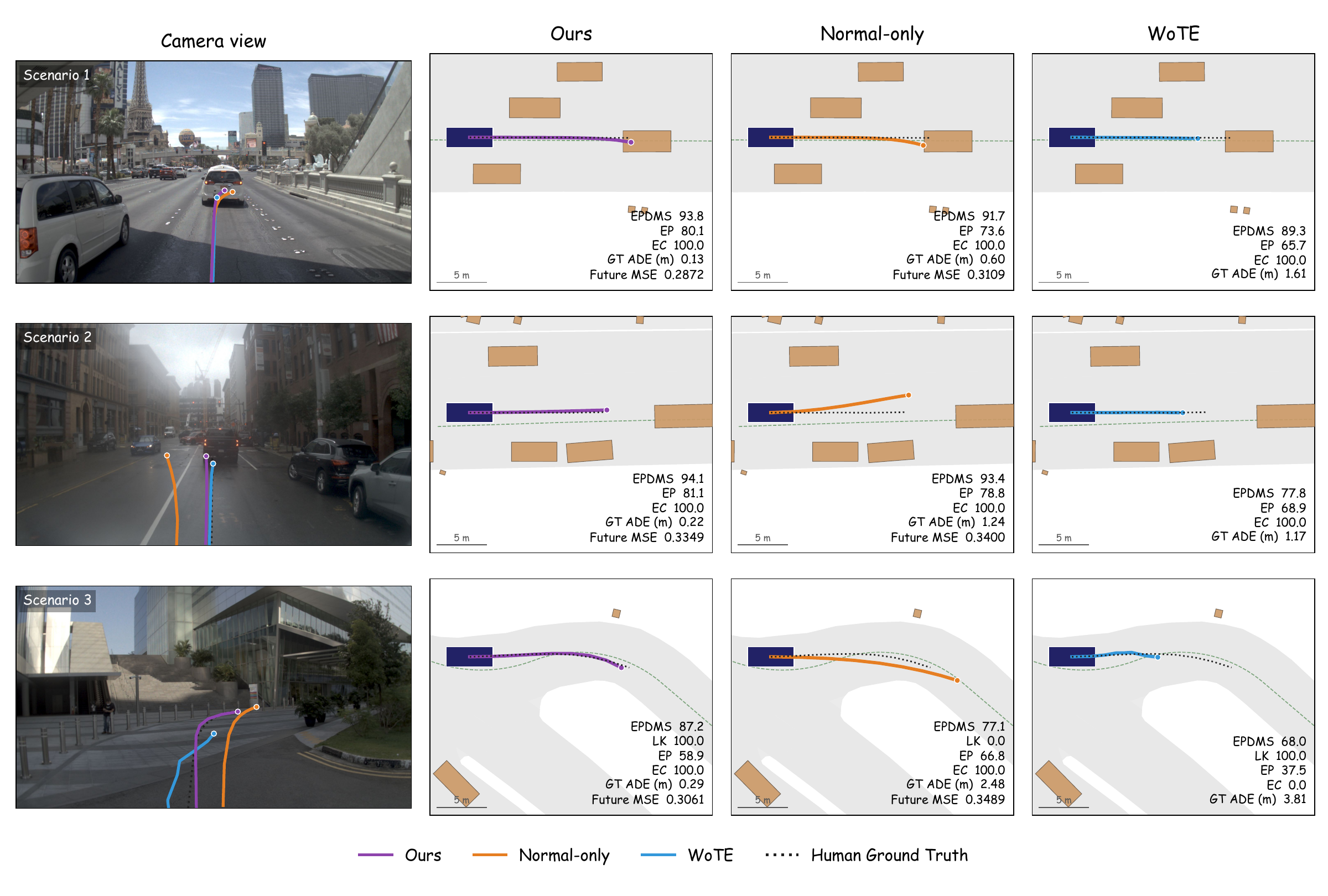}%
    }
    \caption{Selected qualitative planning comparisons using previously evaluated checkpoints. Each row presents one scene, with a camera view followed by Ours, Normal-only, and WoTE. The black dotted line is the human GT trajectory; colored lines show the actual selected trajectories. BEV panels share the same map, current objects, and spatial extent within each row. EPDMS and its displayed sub-scores use a 0--100 scale. GT ADE is the mean position error over eight corresponding future time steps. Future MSE measures expert-trajectory-conditioned future DINO feature prediction for EditWM and Normal.}
    \label{fig:planning_cases}
\end{figure*}

The updated system obtains 91.05 EPDMS, compared with 90.84 for the no-future reference and 90.51 for Normal-only. These correspond to differences of approximately 0.21 and 0.54 points, respectively.

The same-pool Oracle reaches 97.73 EPDMS, leaving a 6.68-point gap between candidate quality and the system's selection. This gap motivates improving the evidence used to distinguish candidates. We next examine future-feature accuracy and its association with planning gains.

\begin{table}[t]
\caption{Future feature prediction under the expert trajectory.}
\label{tab:world_model}
\centering
\small
\setlength{\tabcolsep}{3pt}
\renewcommand{\arraystretch}{1.13}
\begin{tabular}{lrr}
\toprule
Metric & WoTE (Normal) & EditWM \\
\midrule
Validation MSE $\downarrow$ & 0.32490 & \textbf{0.29693} \\
navtest MSE $\downarrow$ & 0.39352 & \textbf{0.37246} \\
navtest cosine $\uparrow$ & 0.95078 & \textbf{0.95355} \\
\bottomrule
\end{tabular}
\par\vspace{3pt}
\parbox{\linewidth}{\footnotesize WoTE denotes our Normal predictor trained on DINOv2 features.}
\end{table}

\subsection{Future Feature Prediction Quality}
\label{sec:prediction_quality}

Table~\ref{tab:world_model} compares Normal and EditWM on the internal validation set and the full navtest set under the expert-trajectory condition described in Sec.~\ref{sec:experimental_setup}. EditWM reduces validation MSE from 0.32490 to 0.29693 (8.61\%) and test MSE from 0.39352 to 0.37246 (5.35\%), while test cosine similarity increases from 0.95078 to 0.95355. The consistent validation and test improvements support the generalization of the residual correction beyond its fitting scenes.

Fig.~\ref{fig:prediction_quality} summarizes the full test set. Panel (a) shows the distribution of per-scene MSE reductions, retaining both positive and negative changes. EditWM reduces MSE in 83.54\% of scenes, with increases in the remaining 16.46\%.

Panel (b) divides scenes into five nearly equal groups according to Normal's MSE. EditWM lowers mean error in every group, with relative reductions of 9.41\%, 8.40\%, 6.86\%, 5.15\%, and 2.06\% from Q1 to Q5. The gains therefore span the baseline-error range, although the relative reduction is smaller in the highest-error group. Panel (c) further shows consistent improvements across the front, rear, left, and right cameras, with relative MSE reductions of 4.95\%, 2.14\%, 6.23\%, and 9.32\%, respectively. Comparisons are made within each view because raw feature scales can differ across cameras.

Together, these results support learning a correction for the feature variation left unexplained by the fixed normal predictor. In an exploratory per-scene comparison of single-model EditWM and Normal-only across all 12,146 test scenes, we group scenes by whether EditWM reduces MSE relative to Normal. Within each group, we count scenes where EditWM strictly improves EPDMS over Normal-only, retaining unchanged and decreased scores in the denominator. The improvement rate is 8.34\% in the reduced-MSE group versus 6.70\% in the remaining group. This 1.63-percentage-point difference (95\% log-level bootstrap confidence interval: 0.14--3.15 points) indicates a higher frequency of planning gains when future prediction improves. Sec.~\ref{sec:qualitative} examines individual planning cases.

\subsection{Qualitative Planning Comparison}
\label{sec:qualitative}

Fig.~\ref{fig:planning_cases} compares Ours, Normal-only, and WoTE~\cite{wote2025} in three navtest scenes. Each BEV overlays the selected and human trajectories with an identical viewing extent within each row. Normal-only omits the EditWM correction and shares our 96-candidate pool and evaluation protocol; WoTE uses its previously evaluated full planner.

\paragraph{Urban following with an adjacent vehicle}
The first scene shows urban car following with a vehicle alongside on the left. Ours closely matches the human path and endpoint (GT ADE: 0.13\,m), while Normal-only bends slightly right near the endpoint (ADE: 0.60\,m) and WoTE makes more conservative forward progress (ADE: 1.61\,m).

\paragraph{Urban following with oncoming traffic}
In the second scene, an oncoming vehicle approaches on the left during urban car following. Ours follows the human trajectory (ADE: 0.22\,m), whereas Normal-only moves forward-left toward oncoming traffic, creating a potential conflict (ADE: 1.24\,m). WoTE again makes more conservative forward progress (ADE: 1.17\,m).

\paragraph{Turning on a narrow urban road}
On a narrow urban bend, Ours follows the human trajectory's curvature (ADE: 0.29\,m; LK=100). Normal-only takes a straighter path toward the road boundary (ADE: 2.48\,m; LK=0), risking roadside contact, while WoTE progresses only partway through the bend (ADE: 3.81\,m). Ours combines lateral alignment with continued forward motion.

\paragraph{Feature prediction and planning scores}
Across the three scenes, expert-conditioned future MSE decreases from 0.3109 to 0.2872, 0.3400 to 0.3349, and 0.3489 to 0.3061. Ours achieves EPDMS of 93.75, 94.09, and 87.16, exceeding both alternatives in every case. In the third scene, it also retains EC=100 where WoTE scores zero. These examples associate more accurate future features with closer human-trajectory agreement and higher planning scores across following, lateral positioning, and turning decisions.

% Balance the final page while allowing the expanded case discussion to flow.
\balance
\section{CONCLUSION}
\label{sec:conclusion}

We presented EditWM, a world model that separates normal scene evolution from unexplained feature changes. Preserving the normal forecast focuses correction learning on these changes and provides refined future representations for candidate-specific trajectory scoring. On NAVSIM navtest, EditWM reduces future-feature MSE by 5.35\% and improves prediction in 83.54\% of scenes; the full system achieves 91.05 EPDMS.

Future work will extend supervision beyond expert trajectories and investigate uncertainty-aware correction in interactive driving, where actions and predictions jointly influence subsequent observations.

\bibliographystyle{IEEEtran}
\bibliography{references}

\end{document}